\documentclass[10pt,journal,cspaper,compsoc]{IEEEtran}

\usepackage[T1]{fontenc}
\usepackage{multirow}
\usepackage{arydshln}

\ifCLASSOPTIONcompsoc
      \else
    \fi

\ifCLASSINFOpdf
  \usepackage[pdftex]{graphicx}
          \else
                  \fi

\usepackage[cmex10]{amsmath}

\ifCLASSOPTIONcompsoc
\usepackage[tight,normalsize,sf,SF]{subfigure}
\else
\usepackage[tight,footnotesize]{subfigure}
\fi

\usepackage{url}

\usepackage{amssymb,amsmath}
\usepackage{algorithm}
\usepackage{algorithmic}
\usepackage{tikz}
\usepackage{color}
\usepackage{booktabs} \usepackage{caption} 

\definecolor{MyDarkBlue}{rgb}{0.85,0.25,0.85} 
\definecolor{MyDarkBlue2}{rgb}{0.15,0.15,0.85} 
\definecolor{MyDarkGreen}{rgb}{0.1,0.6,0.1} 
\definecolor{MyDarkBoh}{rgb}{0.1,0.6,0.8} 
\definecolor{MyDarkpink}{rgb}{0.8,0,0}

\newcommand{\corv}[1]{{#1}}

\newcommand{\DT}[2]{{\color{MyDarkBoh}  #2}}

\def\b{\mathbf{b}}

\newtheorem{theorem}{Theorem}[section]

\newcommand{\qed}{\nobreak \ifvmode \relax \else
      \ifdim\lastskip<1.5em \hskip-\lastskip
      \hskip1.5em plus0em minus0.5em \fi \nobreak
      \vrule height0.75em width0.5em depth0.25em\fi}

\def\argmin{\mathop{\mathrm{argmin}}}   

\def\Pp{\bf \mathcal{B}}
\def\C{{\bf C}}

\def\L{{\bf L}}

\def\tr{\text{Tr}}
\newcommand{\x}{{\bf x}}

\newcommand{\y}{{\bf y}}
\renewcommand{\xi}{{\bf x}_i}
\newcommand{\xsi}{{\bf x}^s_i}
\newcommand{\xti}{{\bf x}^t_i}
\newcommand{\xs}{{\bf x}^s}
\newcommand{\xt}{{\bf x}^t}

\newcommand{\xtj}{{\bf x}^t_j}

\newcommand{\Tmu}{{\bf T}{\#\mu}}
\newcommand{\Tmus}{{\bf T}{\#\mu_s}}
\newcommand{\T}{{\bf T}}
\newcommand{\Tzero}{{\bf T}_{0}}
\newcommand{\Gzero}{{\boldsymbol{\gamma}}_{0}}
\newcommand{\G}{{\boldsymbol{\gamma}}}
\newcommand{\Gzeroreg}{\boldsymbol{ \gamma}^\lambda_{0}}
\newcommand{\Wp}{W_{2}}
\newcommand{\diag}{\text{diag}}

\newcommand{\prob}[1]{\mathcal{P}(#1)}

\newcommand{\proba}[1]{\mathbf{P}(#1)}
\newcommand{\probas}[1]{\mathbf{P}_s(#1)}
\newcommand{\probat}[1]{\mathbf{P}_t(#1)}

\newcommand{\Xt}{{\bf X}_t}
\newcommand{\Xs}{{\bf X}_s}
\newcommand{\Yt}{{\bf Y}_t}
\newcommand{\Ys}{{\bf Y}_s}

\def\R{\mathbb{R}}

\def\C{\mathbf{C}}

\def\X{\mathbf{X}}
\def\x{\mathbf{x}}
\def\A{\mathbf{A}}

\def\y{\mathbf{y}}

\def\ie{{\em i.e.}}
\def\eg{{\em e.g.}}

\DeclareMathOperator*{\E}{\mathbb{E}}

\def\equaldef{\stackrel{\text{def}}{=}}

\newtheorem{proposition*}{\textbf{Proposition}}
\newtheorem{theorem*}{\textbf{Theorem}}
\newtheorem{remark*}{\textbf{Remark}}
\newtheorem{definition*}{\textbf{Definition}}

\begin{document}
\title{Optimal Transport for Domain Adaptation}

\author{Nicolas~Courty, R\'emi~Flamary, Devis~Tuia,~\IEEEmembership{Senior Member,~IEEE}, Alain~Rakotomamonjy,~\IEEEmembership{Member,~IEEE} \thanks{Manuscript received January 2015; revised September 2015.}
}

\markboth{IEEE Transactions on Pattern Analysis and Machine Intelligence,~Vol.~X, No.~X, January~XX}{Courty et al.: XXXXX }

\IEEEcompsoctitleabstractindextext{\begin{abstract}
Domain adaptation is one
of the most challenging tasks of modern data analytics. If the
adaptation is done correctly, models built on a specific data representation
become more robust when confronted to data depicting the same  classes, 
but {described by another observation system}. Among the many strategies proposed, finding domain-invariant representations has shown
excellent properties, in particular since it allows to train a
 unique classifier effective in all domains. In this paper, we propose a regularized unsupervised
optimal transportation model to perform the alignment of the
representations in the source and target domains. We learn a
transportation plan matching both PDFs, which constrains labeled
samples of the same class in the source domain to remain close during transport. This
way, we exploit at the same time the labeled samples  in the
source and the distributions observed in both
domains. Experiments on toy and challenging real visual adaptation
examples show the interest of the method, that consistently outperforms
 state of the art approaches. In addition, numerical experiments
   show that our approach leads to better performances on domain invariant deep learning features and can be easily adapted to
   the semi-supervised case where few labeled samples are available in the target domain.
\end{abstract}

\begin{keywords}
Unsupervised Domain Adaptation, Optimal Transport, Transfer Learning, Visual Adaptation, Classification.
\end{keywords}}

\maketitle

\IEEEdisplaynotcompsoctitleabstractindextext

\IEEEpeerreviewmaketitle

\section{Introduction}

\IEEEPARstart{M}{odern} data analytics are based on the availability of large volumes of data, sensed by a variety of acquisition devices and at high temporal frequency. But this large amounts of heterogeneous data also make the task of learning semantic concepts more difficult, since the data used for learning a decision
function and those used for inference tend not to follow the same distribution.
Discrepancies (also known as drift) in data distribution are due to
several reasons and are application-dependent.  In computer vision,
this problem is known as the visual adaptation domain problem, where domain drifts occur when changing lighting conditions,
acquisition devices, or by considering the presence or absence of
backgrounds. In speech processing, learning from one speaker and
trying to deploy an application targeted to a wide public may also be
hindered by the differences in background noise, tone or gender of the
speaker. In remote sensing image analysis, one would like to leverage
from labels defined over one city image to
classify the land occupation of another city. The drifts observed in
the probability density function (PDF) of remote sensing images are
caused by variety of factors: different corrections for atmospheric
scattering, daylight conditions at the hour of acquisition or even
slight changes in the chemical composition of the materials.

For those reasons, several works have coped with these drift
problems by  developing learning methods able to transfer knowledge from a source domain 
to a target domain for which data have different PDFs. Learning in
this PDF discrepancy context is denoted as the
domain adaptation problem~\cite{Pan10}. In this work, we address the most difficult
variant of this problem, denoted as {\bf unsupervised domain adaptation},
where data labels are only available in the source domain. We tackle this problem by assuming that the effects of the 
drifts can be reduced if data undergo a phase of \emph{adaptation}
(typically, a non-linear mapping) where both domains look more alike.

\begin{figure*}[!t]
 \centering
\includegraphics[width=1\linewidth]{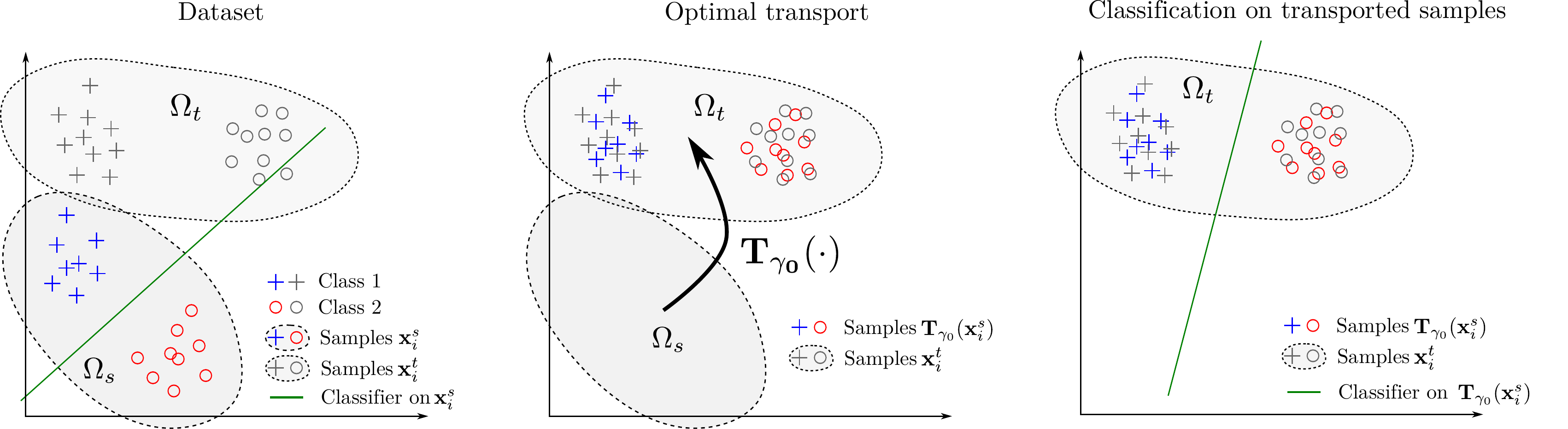}
	\caption{Illustration of the proposed approach for domain
          adaptation. (left) dataset for training, \emph{i.e.} source
          domain, and
          testing, \emph{i.e.} target domain. Note that a classifier
          estimated on
        the training examples clearly does not fit the target
        data. (middle) a data dependent transportation map $\T_{\G0}$ is estimated and
        used to transport the training samples onto the target
        domain. Note that this transformation is usually not linear. 
        (right) the transported labeled samples are used for
        estimating a classifier in the target domain.}
  \label{fig:illus_otda}
\end{figure*}

 Several theoretical works~\cite{bendavid10a,mansour09,germain13} have
 emphasized the role played by the divergence {between} the data
  probability distribution functions of the domains. These works
have  led to a
 principled way of solving the domain adaptation problem:
 transform data so as to make their distributions ``closer'', and
  use the label information available in the source domain to learn
 a classifier {in the transformed domain}, which can be applied
to the target domain. Our work follows the
 same intuition and proposes a
 transformation of the source data  that fits {\bf a least effort
   principle}, \ie~ an effect that is minimal with respect to a
 transformation cost or metric.  In this sense, the adaptation problem
 boils down to: {\em i)} finding a  transformation of the
 input data matching the source and target distributions and
 then {\em ii)} learning a
 new classifier from the transformed source samples.  This process is
 depicted in Figure~\ref{fig:illus_otda}. In this paper, we advocate a solution 
 for finding this transformation based on  \emph{optimal transport}.

Optimal Transport (OT) problems have recently raised interest in
several fields, in particular because OT theory can be used for
computing distances between probability distributions. Those distances,
known under several names in the literature (Wasserstein,
Monge-Kantorovich or Earth Mover distances) have  important properties: {\em i)} {They can be evaluated
  directly on empirical estimates of the distributions without having
  to smoothen the{m} using non-parametric or
  semi-parametric approaches}; {\em ii)} By exploiting the
  geometry of the underlying metric space, they provide meaningful
  distances even when the supports of the distributions do not
  overlap.
Leveraging from th{ese} propert{ies}, we introduce
a novel framework for unsupervised domain adaptation, which
consists in learning an optimal transportation based on empirical
observations.  In addition, we propose several regularization
  terms that favor learning of better transformations  \emph{w.r.t.} the adaptation problem.
They can either encode class information contained in  the  source
  domain or promote the preservation of neighborhood structures.
 {An efficient algorithm is proposed for solving
the resulting regularized optimal transport optimization problem.} 
Finally, this framework can also easily be  extended to the
  semi-supervised case, where few labels are available in the
  target domain, by a simple and elegant modification
in the optimal transport optimization problem.

The remainder of this Section presents related works, while Section~\ref{sec:prob} formalizes the problem of unsupervised domain adaptation and
discusses the use of optimal transport for its resolution.
 Section~\ref{sec:reg} introduces  optimal transport
and its regularized version. Section \ref{sec:transp-regul-doma} presents the  proposed  regularization terms tailored to fit the
domain adaptation constraints. Section~\ref{sec:algo} discusses algorithms for solving the regularized optimal transport problem efficiently.
Section~\ref{sec:numer-exper} evaluates the relevance of our domain adaptation framework through both
synthetic and real-world examples.

\subsection{Related works}

\noindent\textbf{Domain adaptation. } 
Domain adaptation strategies can be roughly divided in two families,
depending on whether they assume the presence of few labels in the
target domain (semi-supervised DA) or not (unsupervised DA).

In the first family, 
  methods which have been proposed include 
 searching for projections that are discriminative in both domains by using inner products between
 source samples and  transformed target
samples~\cite{saenko10,Kul11,Jhuo12}. Learning projections, for
which labeled samples of the target domain fall on the correct side of
a large margin classifier trained on the source data,
have also been proposed~\cite{Hof13}. 
Several works based on  extraction of common features under pairwise
constraints have also been introduced as domain adaptation
strategies~\cite{Ham05,Wang11,Tui15d}.

The second family tackles the domain adaptation problem assuming, as
in this paper, that no labels are available in the target domain. 
Besides works dealing with sample reweighting~\cite{Sug08}, many works
have considered finding a common feature representation for the two
(or more) domains.  Since the representation,  or \emph{latent space},
is common to all domains, projected labeled samples from the source
domain can be used to train a classifier that is general~\cite{Dau07,Pan11}. 
A common strategy is to propose methods that aim at
finding representations in which domains match in some sense. For instance, 
adaptation can be performed by matching the means of the
domains in the feature space~\cite{Pan11}, aligning the domains by
their correlations~\cite{Kum12} or by using pairwise
constraints~\cite{Wan09}. 
In most of these works, feature extraction is the key tool
for finding a common latent space that embeds 
discriminative information shared by all domains. 

Recently, the unsupervised domain adaptation problem has been
  revisited by considering strategies based on a gradual alignment of 
a feature
  representation.  In~\cite{gopalan11}, authors start from the
hypothesis that domain adaptation can be better estimated when
comparing gradual distortions. Therefore, they use intermediary
projections of both domains along the Grassmannian geodesic connecting
the source and target  eigenvectors. In~\cite{gong12,zheng12},
 all sets of transformed intermediary domains are obtained
by using  a geodesic-flow kernel. While these methods
have the advantage of providing easily computable out-of-sample
extensions (by projecting unseen samples onto the latent space
eigenvectors), the transformation defined {remains} global and
is applied in the same way to the whole target domain. An approach combining  sample reweighting logic with  representation transfer is found in~\cite{Zha13}, where authors extend the sample re-weighing to  reproducing kernel Hilbert space through the use of surrogate kernels. The transformation achieved is again a global linear transformation that helps in aligning domains.

Our proposition strongly differs from those reviewed above, as it
defines a local transformation for each sample in the source
domain. In this sense, the domain adaptation problem can be seen as a
graph matching problem~\cite{Luo01,Cae06,Cae09} as each source
sample has to be mapped on target samples  under the constraint
of  marginal distribution preservation.

\noindent\textbf{Optimal Transport and Machine Learning.}
The optimal transport problem has first been introduced by the French mathematician Gaspard Monge in the middle of the 19th century
as a way to find a minimal effort solution to the transport of a given mass of dirt into a given hole. The problem reappeared in the middle of the 20th century in the work of Kantorovitch~\cite{Kantorovich42} and found 
recently surprising new developments as a polyvalent tool for several fundamental problems~\cite{Villani09}{. It was applied in a wide panel of fields, including} computational fluid mechanics~\cite{Benamou00}, 
color transfer between multiple images or morphing in the context of image processing~\cite{Rabin12, ferradans13,BRPP13}, interpolation schemes in computer graphics~\cite{Bonneel11}, and economics{, via matching} and equilibriums problems~\cite{carlier14}. 

Despite the appealing properties {and application success stories}, {the machine learning community has considered optimal transport only recently} (see, for instance, works considering the computation of distances between histograms~\cite{Cuturi13} or label propagation in graphs~\cite{Solomon14}); the main reason being the high computational cost induced by the computation of the optimal transportation plan. However, new computing strategies have emerged~\cite{Cuturi13,Cuturi14,BRPP13} and {made} possible the application of OT distances in operational settings.

\section{{Optimal transport and application to domain adaptation}}
\label{sec:prob}

{In this section, we present the general unsupervised domain adaptation problem and show how it can be addressed from an optimal transport perspective.}

\subsection{{{Problem} and theoretical motivations}}

Let $\Omega\in\R^d$ be an input measurable space of dimension $d$ and $\mathcal{C}$ the set of {possible} labels. $\prob{\Omega}$ denotes the set of all
 probability measures over $\Omega$. The standard learning paradigm
assumes  the existence of a set of training data  $\Xs
= \{ \xsi \}_{i=1}^{N_s}$ associated with a set of class labels $\Ys = \{y_i^s\}_{i=1}^{N_s}$, with $y_i^s \in \mathcal{C}$, and a 
testing set  $\Xt = \{  \xti \}_{i=1}^{N_t}$ with unknown labels.
In order to infer the set of labels $\Yt$  associated with $\Xt$,
one usually relies on an empirical estimate of the joint probability
distribution $\proba{\x,y} \in \prob{\Omega\times\mathcal{C}}$ from
$(\Xs,\Ys)$, and assumes that $\Xs$ and $\Xt$ are drawn from
the same distribution $ \proba{\x} \in \prob{\Omega}$. 

\subsection{Domain adaptation as a transportation problem}
In domain adaptation problems, one assumes the existence of two distinct joint probability distributions $\probas{\xs,y}$ and $\probat{\xt,y}$,
respectively related to  a {\em source} and a {\em
  target} domains, noted  as $\Omega_s$ and $\Omega_t$. In the following,  $\mu_s$ and $\mu_t$ are their respective
marginal distributions over $\X$.  We also denote $f_s$ and $f_t$ the true labeling
functions, \ie\ the Bayes decision functions in each domain. 

{At least o}ne of the  two following assumptions is generally {made} by most {domain adaptation methods}:
\begin{itemize}
\item {\bf Class imbalance}:  Label distributions are different in the two domains ($\probas{y}\neq\probat{y}$), but the conditional distributions of the samples with respect to the labels are the same ($\probas{\xs|y}=\probat{\xt|y}$);
\item {\bf Covariate shift}:  Conditional distributions of the
  labels with respect to the data are equal
  ($\probas{y|\xs}=\probat{y|\xt}$, or equivalently $f_s=f_t=f$). However, data distributions in the two domains are supposed to be different
  ($\probas{\xs}\neq\probat{\xt}$). For the
  adaptation techniques to be effective, this difference needs to be small~\cite{bendavid10a}. 
\end{itemize}
{I}n real world
applications, the drift occurring between the source and the target
domains generally implies a change in both marginal and conditional
distributions.

\begin{figure*}[!t]
  \centering
	\subfigure[]{\includegraphics[width=0.5\linewidth,height=3.7cm]{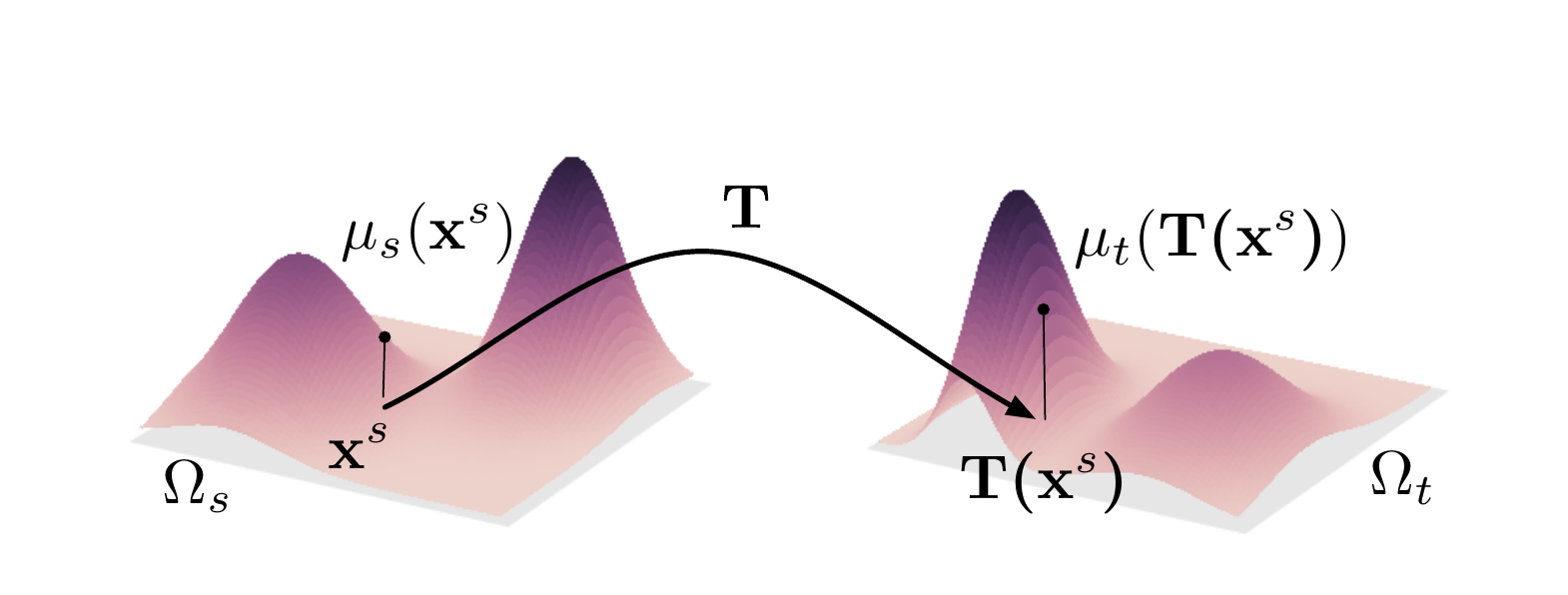}}  
	\subfigure[]{\includegraphics[width=0.23\linewidth,height=4cm]{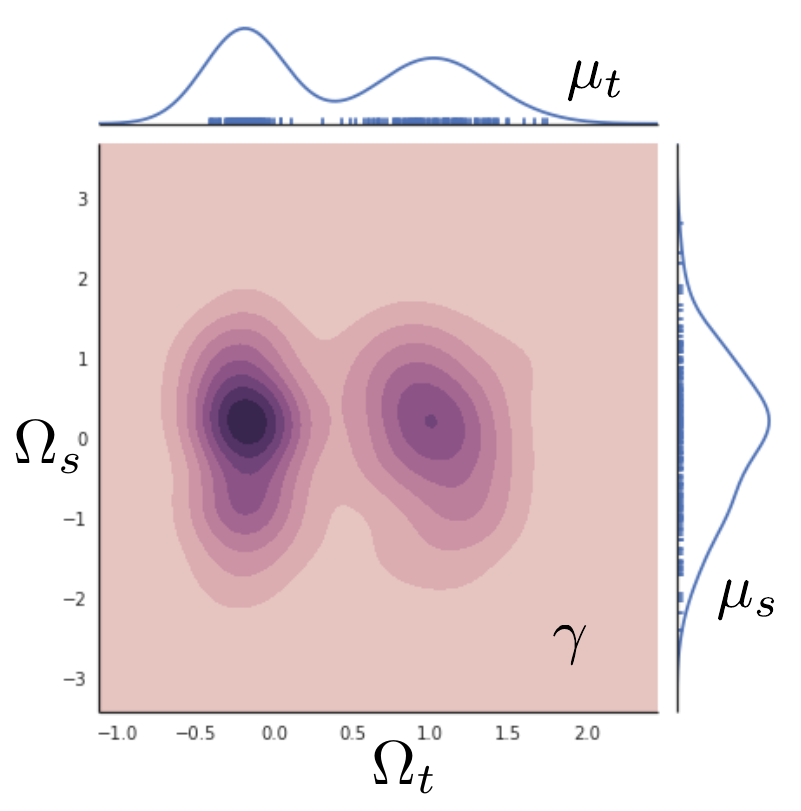}}  
	\subfigure[]{\includegraphics[width=0.25\linewidth,height=4cm]{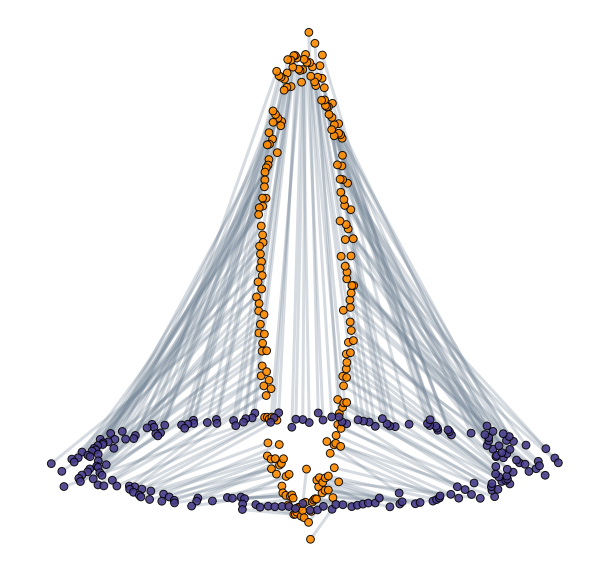}}  
	\caption{Illustration of the optimal transport problem. (a) Monge problem over 2D domains. $\T$ is a push-forward from $\Omega_s$ to $\Omega_t$. (b) Kantorovich relaxation over 1D domains: $\G$ can be seen as a joint probability distribution with marginals $\mu_s$ and $\mu_t$. (c) Illustration of the solution of the Kantorovich relaxation computed between two ellipsoidal distributions in 2D. The grey line between two points indicate a non-zero coupling between them.}
  \label{fig:illus_transp}
\end{figure*}

In our work, we assume that the
domain drift is due to an unknown, possibly nonlinear transformation of the input space
$\bf T:\Omega_s \rightarrow \Omega_t$. This transformation may have a
physical interpretation (e.g. change in {the} acquisition conditions,
sensor drifts, thermal noise, etc.). It can also be directly
caused  by the unknown process that generates the data. 
 \corv{Additionnally,
  we also suppose that the transformation preserves the conditional
  distribution, \ie {
$$\probas{y|\xs}=\probat{y|{\bf T}(\xs)}.$$} This means that the label
information is preserved by the
transformation, and the Bayes decision functions are tied
through the equation $f_t({\bf T}(\x))=f_s(\x)$.  

Another insight can be provided regarding the transformation $\bf T$. From a probabilistic
point of view, $\bf T$ transforms the measure $\mu$ in its
{\em image measure}, noted $\Tmu$, which is  another probability measure over
$\Omega_t$ satisfying
\begin{equation}
 \Tmu(\x) = \mu(\T^{-1}(\x)),\;\;\;   \forall \x \in \Omega_t
\end{equation}
 $\T$ is said to be a {\bf transport map} or {\bf push-forward} from $\mu_s$ {to} $\mu_t$  if $\Tmus=\mu_t$ ({as}
illustrated in Figure~\ref{fig:illus_transp}.a). Under this
assumption, $\Xt$ are drawn from the same PDF as $\Tmu_s$. This provides a
principled way to solve the adaptation problem:
\begin{enumerate}
\item {Estimate $\mu_s$ and $\mu_t$ from $\Xs$ and $\Xt$} (Equation \eqref{eq:mus})
\item {Find a transport  map $\bf T$ from  $\mu_s$ to $\mu_t$}
\item \corv{Use $\bf T$ to transport labeled samples $\X_s$ and
        { train a classifier from them}.}\end{enumerate}
} 

Searching for $\bf T$ in the space of all possible transformations
is intractable, and some restrictions need to be imposed.
Here, {w}e propose that  {\bf T} should be
chosen so as to minimize a transportation cost $C(\T)$ expressed as: 
\begin{equation}
C(\T) = \int_{\Omega_s} c(\x,\T(\x))d\mu(\x),
\end{equation}
where the cost function $c:\Omega_s \times \Omega_t \rightarrow
\mathbb{R}^+$ is a  distance function over the metric
space $\Omega${. $C(\T)$ can be interpreted} as the energy required to move a
probability mass $\mu(\x)$ from $\x$ to $\T(\x)$.

{The problem of finding such a transportation of minimal cost has
already been investigated in the literature. For instance, 
the optimal transportation problem as defined by Monge is the}  solution of the following minimization problem:
\begin{equation}
 \Tzero = \argmin_\T  \int_{\Omega_s} c(\x,\T(\x))d\mu(\x), \;\;\;\;\; \text{s.t.  } \; \Tmus=\mu_t
\end{equation}
 The  Kantorovitch formulation of the optimal transport{ation}~\cite{Kantorovich42} is a convex relaxation of the above Monge problem. 
Indeed, let us define $\Pi$ as the set of all  probabilistic couplings $\in \mathcal{P}(\Omega_s \times \Omega_t)$ with marginals $\mu_s$ and $\mu_t$. The Kantorovitch
problem  seeks for a general coupling $\G \in\Pi$ between
$\Omega_s$ and $\Omega_t$: 

\begin{equation}
 \Gzero = \argmin_{\G \in \Pi}  \int_{\Omega_s\times\Omega_t} c(\xs,\xt)d\G(\xs,\xt)
 \label{eq:kanto}
\end{equation}
In this formulation, $\G$ can be understood as a joint probability
measure with marginals $\mu_s$ and $\mu_t$ as depicted in
Figure~\ref{fig:illus_transp}.b. 
$\Gzero$ is also known as \textbf{transportation plan}
\cite{santambrogio2015optimal}. It allows to define the {\bf Wasserstein distance} of order $p$ between $\mu_s$ and $\mu_t$. This distance is formalized as
\begin{eqnarray}
 W_p(\mu_s,\mu_t) & \equaldef& \left(\inf_{\G \in \Pi} \int_{\Omega_s\times\Omega_t} d(\xs,\xt)^p d\G(\xs,\xt)\right)^{\frac{1}{p}}\nonumber\\
 &=&  \inf_{\G \in \Pi}  \left\{ \left(\E_{\xs\sim\mu_s,\xt\sim\mu_t} d(\xs,\xt)^p\right)^{\frac{1}{p}}\right\}  
 \label{eq:wasserstein}
\end{eqnarray} 
where $d$ is a distance and the corresponding cost function $c(\xs,\xt)=d(\xs,\xt)^p$.
The Wasserstein distance is  also known as the Earth Mover Distance in the computer
vision community~\cite{rubner98} and it defines a metric over the space of
integrable squared probability measure{s}.

In the remainder, we consider the squared $\ell_2$ Euclidean distance as a
cost function, $c(\x,\y)=\|\x-\y\|_2^2$ for computing optimal
transportation.  As a consequence, we evaluate
distances between measures according to the squared Wasserstein distance
$W^2_2$ associated with the Euclidean distance $d(\x,\y)=\|\x-\y\|_2$.
The main rationale for this choice is that  it experimentally provided 
the best result on average (as shown in the   supplementary material).
Nevertheless, other cost functions better suited
 to the nature of specific data can be considered, depending on the
 application at hand and the data representation, as discussed
   more in details in Section \ref{sec:strength-limits-opti}.

\section{Regularized discrete optimal transport}
\label{sec:reg}
{T}his section discusses the problem of
optimal transport for domain adaptation.  In the first
part, we introduce the OT optimization problem on discrete empirical distributions. 
{Then, we discuss a regularized variant of this discrete optimal transport problem. Finally, we address 
the question of how the resulting probabilistic coupling can be
used for mapping samples from source to target domain. }

\subsection{Discrete optimal transport}

When $\mu_s$ and $\mu_t$ are only accessible through discrete samples, the corresponding empirical 
distributions can be written as
\begin{equation}
\mu_s = \sum_{i=1}^{n_s} p^s_{i}  \delta_{\xsi}, \;\;\; \mu_t = \sum_{i=1}^{n_t} p^t_{i} \delta_{\xti} 
\label{eq:mus}
\end{equation}
where $\delta_{\xi}$ is the Dirac function at location $\xi \in
\mathbb{R}^d$. $p^s_{i}$ and $p^t_{i}$ are probability masses
associated to the $i$-th sample and belong to the probability
simplex, {\em i.e.} $\sum_{i=1}^{n_s}  p^s_{i} = \sum_{i=1}^{n_t}
p^t_{i} = 1$.
It is  straightforward to adapt the Kantorovich formulation of optimal transport problem to the discrete case.
We denote $\Pp$ the set of  probabilistic coupling{s} between the two empirical distributions defined as:  
\begin{eqnarray}
\label{eq:coupling}
\Pp  = \left\{ \G \in (\mathbb{R}^+)^{n_s \times n_t} | \:  \G \mathbf{1}_{n_t} = \mu_s,\G^T \mathbf{1}_{n_s} = \mu_t \right\}
\end{eqnarray}
where $ \mathbf{1}_{d}$ is a $d$-dimensional vector of ones. The Kantorovitch formulation of the optimal transport~\cite{Kantorovich42} reads:
\begin{equation}
\label{eq:gzero}
\Gzero = \argmin_{\G \in \Pp}\quad \left < \G, \C\right >_F
\end{equation}
where $\left< .,. \right>_F$ is the Frobenius dot product and $\C \geq
0$ is the cost function matrix{, whose} term $C(i,j)=c(\xsi,\xtj)$
denotes the cost to move a probability mass from $\xsi$ to $\xtj$. {As 
previously detailed}, this cost was chosen as the squared Euclidean distance between the two locations,
i.e. $C(i,j) = ||\xsi - \xtj||_2^2$.

Note that when $n_s=n_t=n$ and $\forall i,j\;\;p^s_{i}=p^t_{j}=1/n$,   $\Gzero$ is simply a permutation matrix. In this case, the optimal transport problem boils down to an optimal assignment problem.
In the general case, it can be shown that $\Gzero$ is a sparse matrix
with at most $n_s+n_t-1$ non zero entries, equating the rank of the constraint  matrix expressing the two marginal constraints.

{P}roblem~\eqref{eq:gzero} is a linear program and
  can be solved with combinatorial algorithms such as the simplex
methods and its network variants (successive shortest path algorithms, Hungarian or relaxation algorithms). 
Yet, the computational complexity was shown to be $O((n_s+n_t) n_s n_t log(n_s+n_t))$
~\cite[p. 472, Th. 12.2]{Ahuja1993} at best, which dampens the
utility of the method when handling large datasets. However, the regularization scheme recently proposed by Cuturi~\cite{Cuturi13} presented
in the next section, allows a very fast computation of
a {transportation plan}.

\subsection{Regularized optimal transport}
\label{sec:regul}
Regularization is a
classical approach used for preventing overfitting when  few samples
are available for learning. 
It can also be used for inducing some properties on 
the solution. In the following, we
discuss a  regularization term recently introduced
for optimal transport problem.

Cuturi~\cite{Cuturi13}  proposed to regularize the
expression of the optimal transport problem  by the entropy of the probabilistic
coupling. The resulting {information-theoretic} regularized version of the
transport $\Gzeroreg$ is the solution of the minimization
problem:
\begin{equation}
\Gzeroreg = \argmin_{\G \in \Pp} \left < \G, \C \right >_F + {\lambda} \Omega_s(\G),
\label{eq:cuturi}
\end{equation}
where $\Omega_s(\G)=\sum_{i,j} \G(i,j) \log \G(i,j)$ computes the negentropy of $\G$. The intuition behind this form of regularization is the following: since most elements
of $\Gzero$ should be zero with high probability, one can look for a
smoother version of the transport, thus lowering its sparsity, by increasing its entropy. 
As a result, the optimal transport $\Gzeroreg$ will have a denser
coupling between the distributions. 
 $\Omega_s(\cdot)$ can also
  be interpreted as a Kullback-Leibler divergence $KL(\G\|\G_u)$ between the joint   probability $\G$ and a uniform joint  probability 
  $\gamma_u(i,j)=\frac{1}{n_sn_t}$. Indeed, by expanding this KL divergence, we have
 $KL(\G\|\G_u)=\log n_sn_t+\sum_{i,j} \G(i,j) \log \G(i,j)$.
The first term is a constant \emph{w.r.t.} $\G$, which means
that we can equivalently use $KL(\G\|\G_u)$ or $\Omega_s(\G)=\sum_{i,j} \G(i,j) \log \G(i,j)$ in Equation \eqref{eq:cuturi}.

Hence, as the parameter $\lambda$ weighting the entropy-based regularization
increases, the  sparsity of $\Gzeroreg$ decreases
and  source points tend to distribute their probability
masses toward more target points.  When
  $\lambda$ becomes very large ($\lambda\rightarrow\infty$), the OT
solution of Equation \eqref{eq:cuturi}
  converges toward
$\gamma_0^\lambda(i,j)\rightarrow\frac{1}{n_sn_t},\forall
i,j$.

Another appealing outcome of  the
regularized OT formulation given in Equation \eqref{eq:cuturi}  is the
derivation of a computationally 
efficient algorithm based on Sinkhorn-Knopp's scaling matrix
approach~\cite{knight08}. This efficient algorithm will also be a key
element in our methodology presented in Section~\ref{sec:transp-regul-doma}.

\subsection{OT-based mapping of the samples}
\label{sec:mapping-samples}

{In the context of domain adaptation, once the probabilistic coupling} $\Gzero$ has been computed,  source samples have to
be transported in the target domain. For this purpose, one can
interpolate the two distributions $\mu_s$ and $\mu_t$ by following the geodesics of the
Wasserstein metric~\cite[Chapter 7]{Villani09}, parameterized by $t
\,\in [0,1]$.  This defines a new distribution
$\hat{\mu}$  such that:
\begin{equation}\label{eq:interp_wass}
\hat{\mu} = \argmin_\mu\quad (1-t)\Wp(\mu_s,\mu)^2 + t \Wp(\mu_t,\mu)^2.
\end{equation}
Still following Villani's book, one can show that for a squared $\ell_2$ cost,  this distribution
boils down to:
\begin{equation}
\hat{\mu} = \sum_{i,j} \Gzero(i,j) \delta_{(1-t)\xsi+ t \xtj}.\label{eq:interp}
\end{equation}
{
Since our goal is to transport the source samples onto the
  target distribution,  we are mainly interested
in the case $t=1$. For this value of $t$,
the novel distribution $\hat \mu$ is a distribution with the same support
of $\mu_t$, since Equation \eqref{eq:interp}
reduces to 
\begin{equation}
\hat{\mu} = \sum_{j} \hat p^t_{j} \delta_{\xtj}.\label{eq:interp2}
\end{equation}
with  $\hat p^t_{j}=\sum_i \Gzero(i,j)$. The weights
$\hat p^t_{j}$ can be seen as the sum of probability  mass coming from
all samples $\{\xsi\}$ that is transferred to sample $\xtj$.  
Alternatively, $\Gzero(i,j) $ also tells
us how much probability mass of $\xsi$ is transferred
to $\xtj$. 
We can exploit this information to compute a transformation
of the source samples.
This transformation can be conveniently expressed with respect to
the target samples as the following  \textbf{barycentric mapping}: 
  \begin{equation}
    \label{eq:bary_map}
    \widehat \xsi= \argmin_{\x \in \R^d}\quad \sum_j  \Gzero(i,j) c(\x,\xtj).
  \end{equation}
where $\xs_i$ is a  given source sample and $ \widehat \xsi$ is its
corresponding image.}
When the cost function is the  squared $\ell_2$ distance, this barycenter corresponds to a
weighted average and the sample is mapped into the convex hull of the
target samples.
 For all source samples, this barycentric mapping can {therefore} be expressed as:
\begin{equation}
\hat{\bf X}_s = {\bf T}_{\Gzero}({\bf X}_s) =  \diag(\Gzero \mathbf{1}_{n_t})^{-1}\Gzero {\bf X}_t.
\label{eq:OTatt1}
\end{equation}
 
The inverse mapping from the target to the source domain can also be easily computed from $\Gzero^T$.
 {Interestingly, one can show~\cite[Eq. 8]{Cuturi14} that this transformation is a first order approximation
  of the true $n_s$ Wasserstein barycenters of the target
  distributions}.
 Also note that when  marginals $\mu_s$ and $\mu_t$ are
uniform, one can easily derive the barycentric mapping as a linear {expression}: 
\begin{equation}
\hat{\bf X}_s =  n_s\Gzero \X_t\quad\text{ and }\quad\hat\X_t =  n_t\Gzero^\top \X_s 
\label{eq:OTattunif}
\end{equation}
for the source and target samples.     
Finally, remark that if $\gamma_0(i,j)=\frac{1}{n_sn_t},\forall
i,j$, then each transported source point converges toward the
center of mass of the target distribution that is $\frac{1}{n_t}\sum_j \xtj$.
This occurs when $\lambda \rightarrow \infty$ in Equation (\ref{eq:cuturi}).

\vspace{-0.3cm}

\subsection{Discussing optimal transport for domain adaptation}
\label{sec:strength-limits-opti}
We discuss here the requirements and conditions of applicability of the proposed method.

\noindent\textbf{Guarantees of recovery of the correct transformation.}
{Our goal for achieving domain adaptation is to
uncover the transformation that occurred between source
and target distributions. 
}
While the family of transformation
that an OT formulation can recover is wide, we provide a  proof
that, for some simple affine transformations of discrete distributions, 
our OT solution is able to  match source and target examples exactly.

  \begin{theorem}
  Let $\mu^s$ and $\mu^t$ be two discrete
distributions with $n$ Diracs  as defined in Equation \eqref{eq:mus}. 
If the following conditions hold
\begin{enumerate}
\item The source samples in $\mu^s$ are
$\x_i^s\in\R^d, \forall i \in 1,\dots,n$ such that $\x_i^s\neq\x_j^s$ if
$i\neq j$ .
\item All weights in the source and
target distributions are $\frac{1}{n}$.
\item The target samples are defined as $\x^t_i=\A\x_i^s+\b$ \ie\ an
  affine tranformation of the source samples.
\item $\b\in\R^d$ and $\A\in\mathcal{S}^+$  is a strictly positive
  definite matrix.
\item The cost function is $c(\x^s,\x^t)=\|\x^s-\x^t\|_2^2$.
\end{enumerate}
then the solution $\Tzero$ of the  optimal transport problem
{(\ref{eq:gzero})} is so that $\Tzero(\x_i^s)=\A\x_i^s+\b=\x_i^t\quad \forall i \in 1,\dots,n$. 
  \end{theorem}

In this case, we retrieve the exact affine transformation 
on the discrete samples, which means that the label information are fully
preserved during transportation. Therefore,  one can {train} a
classifier on the mapped samples with no generalization loss.
We provide a simple demonstration in the supplementary material.\\

\noindent\textbf{Choosing the cost function.}
{In this work, we have mainly considered a $\ell_2$-based
cost function. Let us now discuss the implication 
of using  a different cost function in our framework}. A
number of norm-based distances have been investigated by mathematicians
 \cite[p 972]{Villani09}. Other types of metrics can also be
 considered, such as Riemannian distances over a manifold~\cite[Part II]{Villani09}, or learnt  metrics~\cite{Cuturi2014}.
Concave cost functions  are also of particular use in real
life problems \cite{gangbo1996geometry}.
Each different cost function  will lead to a
different OT plan $\Gzero$, but the cost itself does not impact the OT optimization
problem, \emph{i.e.} the solver is independent from the cost
function. Nonetheless,  since $c(\cdot,\cdot)$
defines the Wasserstein geodesic, the interpolation
between domains defined in Equation \eqref{eq:interp_wass} leads to a
different trajectory (potentially non-unique). 
Equation
\eqref{eq:interp}, which  corresponds to $c(\cdot,\cdot)$, is a squared $\ell_2$ distance, so it does not hold
anymore. Nevertheless, the solution of \eqref{eq:interp_wass} for $t=1$
does not depend on the cost $c$ and one can still use the proposed barycentric
mapping \eqref{eq:bary_map}. For instance if the cost function is based on the $\ell_1$ norm, the
transported samples will be estimated using a component-wise weighted median.
Unfortunately,  for more complex cost functions, the barycentric mapping might be complex
to estimate.

\section{Class-regularization for domain adaptation }
\label{sec:transp-regul-doma}

In this section we explore regularization terms that preserve label information and sample neighborhood  during transportation.
Finally, we discuss the semi-supervised case and show that label
  information in the target domain can be  effectively included  in  he proposed  model.

\subsection{Regularizing the transport with class labels}

Optimal transport, as it has been  presented in the previous section, does not
use any class information. 
 However, and even if  our goal is unsupervised domain
adaptation, class labels are available in the source domain.
 This information is  typically used only during
the decision function learning stage, which
follows the adaptation step.
Our  proposition {is} to take advantage of the label information for estimating a better transport. More precisely,
we aim at  penalizing
couplings that match source samples with different labels
to same target samples. 

To this end, we propose to add a new term to the regularized optimal
transport, leading to the following optimization problem:
\begin{equation}
\min_{\G \in \Pp}\quad \left < \G, \C \right >_F +\lambda
\Omega_s(\G) +\eta \Omega_c(\G),
\label{eq:optreg_disc}
\end{equation}
where $\eta\geq 0$  and $\Omega_c(\cdot)$ is a class-based
regularization term.

In this work, we propose and study two choices for this regularizer $\Omega_c(\cdot)$. The first is based on group sparsity and
promote{s a  probabilistic coupling $\Gzero$ where
a given target sample receives masses from  source samples which
have same labels.
 The second is based on graph Laplacian regularization
and promote{s} a locally smooth and class{-regular} structure in the source transported samples.

\subsubsection{Regularization with group-{sparsity}}
\label{sec:class-regul-with}

With the first regularizer, our objective is to exploit label
information in the optimal transport computation. We suppose
 that all  samples in the source domain have labels.
The main intuition underlying the use of this group-sparse regularizer
is that we would like each target sample to receive masses only from
source samples that have the same label.
As a consequence, we expect that a given target sample will be involved
in the representation of transported source samples
as defined in Equation (\ref{eq:OTatt1}), but only for
samples from the source domain of the same class.
This behaviour can be induced by means of a group-sparse penalty
on the columns of $\G$.

This
approach has been introduced in our preliminary
work~\cite{courty2014domain}. In that paper, we proposed a $\ell_p-\ell_1$ regularization term with $p<1$ (mainly for
algorithmic reasons).  When applying a majoration-minimization
technique on the $\ell_p-\ell_1$ norm, the problem can be cast as problem
\eqref{eq:cuturi} and can be solved using the efficient Sinkhorn-Knopp
algorithm at each iteration. However, this regularization term with 
$p<1$ is non-convex and thus the proposed algorithm is guaranteed
to converge only to  local stationary points.

In this paper, we retain the convexity of the underlying problem and use the
  convex group-lasso regularizer $\ell_1-\ell_2$ instead. This regularizer  is defined as 
 \begin{equation}
\Omega_c(\G) =\sum_j \sum_{cl} || \G(\mathcal{I}_{cl},j)||_2,
\label{eq:cuturigroup}
\end{equation}
where  $||\cdot||_2$ denotes the $\ell_2$ norm and $\mathcal{I}_{cl}$ contains the {indices} of rows in $\G$ related to
 source domain samples of  class ${cl}$. Hence, $ \G(\mathcal{I}_{cl},j)$ is a vector  containing coefficients of the $j$th column of $\G$
associated to class ${cl}$. Since the  $j$th column of $\G$ is related to the $j$th target sample,
this regularizer will induce the desired sparse representation in the target sample. 
Among other benefits, the convexity of the corresponding problem allows to use an efficient generic optimization scheme, 
presented in Section \ref{sec:algor-solv-class}.

{
Ideally,  with this regularizer we expect that the
masses corresponding to each group of labels 
are matching samples of the source and
target domains exclusively. Hence, for the domain adaptation
problem to have a relevant solution, 
the distributions of labels are expected to be preserved in both the source and target distributions. We thus need to have $\probas{y}=\probat{y}$.
This assumption, which is a classical
assumption in the field of learning, is nevertheless a mild
requirement since, in practice, small deviations of proportions do not prevent the method from working (see reference~\cite{tuia15} for experimental results on this particular issue).

}

\subsubsection{Laplacian regularization}
\label{sec:class-regul-using}

This regularization term {aims at preserving the data structure
  -- approximated by a graph -- during  transport \cite{ferradans13,carreira14}.} 
Intuitively, we would like  similar samples in 
the source domain to also be similar after transportation. 
Hence, denote as $\hat\x^s_i$ the
transported source sample  $\x^s_i$, with $\hat\x^s_i$ being linearly dependent on the transportation matrix $\G$ through Equation \eqref{eq:OTatt1}.
Now, given a positive symmetric similarity matrix $\mathbf{S_s}$ of samples
in the source domain, our regularization term is defined as
\begin{equation}
  \label{eq:reglapsurce}
  \Omega_c(\G)=\frac{1}{N_s^2} \sum_{i,j}S_s({i,j})\|\hat\x^s_i-\hat\x^s_j\|_2^2,
\end{equation}
where $S_s({i,j})\geq 0$ are the coefficients of matrix
$\mathbf{S}_s\in\R^{N_s\times N_s}$
that encode{s}   similarity between pairs of source sample. 
In  order to further preserve class structures,
we can sparsify similarities for samples of different
classes. In practice, we thus impose $S_s(i,j)=0$ if $y_i^s\neq
y_j^s$.

The above equation can be simplified when the marginal distributions are uniform. In that case, transported source samples
can be computed according to Equation (\ref{eq:OTattunif}).
Hence, $\Omega_c(\G)$ boils down to 
\begin{align}
\Omega_c(\G)=\tr(\X_t^\top\G^\top \L_s \G\X_t),\label{eq:reglaps}
\end{align}
where $\mathbf{L}_s=\text{diag}(\mathbf{S}_s {\mathbf 1}) - \mathbf{S}_s$ is
the Laplacian of the graph  $\mathbf{S}_s$. {The regularizer is therefore}  quadratic \emph{w.r.t.} $\G$.

The regularization terms \eqref{eq:reglapsurce} or 
\eqref{eq:reglaps} 
 are defined
based on the transported source samples.  When a similarity
information is also available  in the target samples,
for instance, through a similarity  matrix  $ \mathbf{S}_t$,
we can take advantage of this knowledge and
 {a} symmetric
Laplacian regularization of the form
\begin{align}
\Omega_c(\G)=(1-\alpha)\tr(\X_t^\top\G^\top \L_s \G\X_t)+\alpha\tr( 
\X_s^\top\G \L_t \G^\top \X_s )\label{eq:reglaps2}\end{align}
{can be used instead. In the above equation}
$\mathbf{L}_t=\text{diag}(\mathbf{S}_t {\mathbf 1}) - \mathbf{S}_t$
{is the Laplacian of the graph in the target domain} and
$0\leq \alpha\leq 1$ is a trade-off parameter
that weights the importance of each part of the regularization
term. 
Note that, unlike the matrix $\mathbf{S}_s$, the similarity matrix $\mathbf{S}_t$ cannot be sparsified according to the class structure, since labels are generally not available for the
target domain.

A regularization term similar 
to   $\Omega_c(\G)$  has been proposed in
 \cite{ferradans13} for histogram adaptation between images.  
However, the authors focused on displacements ($\hat\x^s_i-
\x^s_i$) instead of on preserving the class structure
of the transported samples.

\subsection{Regularizing for semi-supervised domain adaptation  }
In semi-supervised domain
  adaptation, few labelled samples are available in the target domain~\cite{Wan11}. Again, such an important information can be exploited by means
of a novel regularization term to be integrated in the original optimal
transport formulation. This regularization term is designed such that
samples in the target domain should only 
be matched with samples in the source domain that have the same labels. {It}  {can be} expressed {as}:
\begin{equation}
\Omega_{semi}(\G) = \langle \G, {\bf M} \rangle 
\label{eq:ssl}
\end{equation}
where $\bf M$ is a $n_s\times n_t$ cost matrix, with ${\bf M}(i,j) = 0$ whenever ${\bf y}^s_i={\bf y}^t_j$ (or $j$ is a sample with unknown label) and $+\infty$ otherwise.  This term has the 
benefit to be parameter free. It 
{boils down} to chang{ing} the original cost  function $\C$, defined in Equation (\ref{eq:gzero}), by adding an infinite cost to undesired matches. Smooth versions of this regularization can be devised, 
for instance, by using
a probabilistic confidence of target sample $\xtj$ to belong to class ${\bf
  y}^t_j$. Though appealing, we have not explored this {latter} option
in this work. It is also noticeable that 
the Laplacian strategy in Equation~\eqref{eq:reglaps2} can also
leverage on these class labels in the target domain through
the definition of matrix $\mathbf{S}_t$ .

 \section{Generalized conditional gradient for solving  regularized OT problems}
 \label{sec:algor-solv-class}
 \label{sec:algo}

In this section, we discuss an efficient algorithm for solving 
optimization problem \eqref{eq:optreg_disc}, that can be used with any of the proposed regularizers. 

Firstly, we characterize the existence of a solution to the problem. 
We remark that regularizers 
given in Equations (\ref{eq:cuturigroup})  and
 \eqref{eq:reglapsurce}
are continuous, thus the
objective function  is continuous. Moreover, since the  constraint set $\Pp$ is a convex, closed and bounded (hence compact) subset of $\R^d$, the
objective function reaches its minimum on $\Pp$.
In addition, if the regularizer is
strictly convex that minimum is unique. This occurs
for instance, for the Laplacian regularization in
Equation  \eqref{eq:reglapsurce}.

Now, let us discuss algorithms for computing optimal transport solution of 
problem \eqref{eq:optreg_disc}. For solving a similar
problem with a Laplacian
regularization term, Ferradans et al. \cite{ferradans13}  used a conditional gradient (CG) algorithm \cite{bertsekas1999nonlinear}. 
This  approach is appealing and could be extended to our problem.
It is an iterative scheme that guarantees  any iterate to belong to
$\Pp$, meaning that any of those iterates is a transportation plan.
{At} each of these iterations, in order to find a feasible search direction, a CG algorithm looks for a minimizer of 
the objective function's linear approximation .
Hence, at each iteration it solves a Linear Program (LP) that is
presumably easier
to handle than the original regularized optimal transport problem.
 Nevertheless, and despite existence of efficient LP solvers such as CPLEX or MOSEK, the
dimensionality of the LP problem makes this LP problem hardly tractable, since it involves  $n_s\times n_t$  variables.

In this work, we aim for {a more scalable algorithm}. To this end, we consider an approach based on a generalization of 
the conditional gradient algorithm~\cite{bredies2009generalized}  denoted as generalized conditional gradient (GCG).

The  framework of the GCG algorithm addresses the general case of constrained minimization of composite functions  defined as
\begin{equation}
  \label{eq:optfws}
  \min_{\G \in \Pp}\quad f(\G) +g(\G),
\end{equation}
where $f(\cdot)$ is a differentiable and possibly non-convex function; $g(\cdot)$ is a convex, possibly non-differentiable function;  $\Pp$ denotes any convex and compact subset of $\R^n$.
As illustrated in Algorithm \ref{algo:fws}, all the  steps of the  GCG 
algorithm are exactly the same as those used for CG,  
except for the search direction part (Line 3). 
The difference is that GCG linearizes only part $f(\cdot)$ of the
composite objective function, instead of the full objective function. This approach is justified when the resulting nonlinear optimization problem can be efficiently solved.
The GCG algorithm
    has been shown by Bredies et al.~\cite{bredies2005equivalence} to converge towards a stationary
    point of Problem \eqref{eq:optfws}. In our case, since $g(\G)$ is
    differentiable,  stronger  convergence results can be provided (see supplementary
      material for a discussion on convergence rate and duality gap monitoring).

\begin{algorithm}[t]
  \caption{Generalized Conditional Gradient}
\label{algo:fws}
\begin{algorithmic}[1]
\STATE Initialize $k=0$ and $\G^0\in\mathcal{P}$ 
\REPEAT
\STATE With $\mathbf{G}\in\nabla f(\G^k)$, solve 
    \begin{equation*}
      \G^\star=\argmin_{\G \in \Pp} \quad\left < \G, \mathbf{G}\right >_F+ g(\G)
    \end{equation*}
\STATE Find the optimal step $\alpha^k$   
    \begin{equation*}
      \alpha^k=\argmin_{0\leq \alpha \leq 1} f(\G^k+\alpha \Delta\G)+g(\G^k+\alpha \Delta\G)
    \end{equation*}
with $\Delta\G=\G^*-\G^k$ 

\STATE $\G^{k+1}\leftarrow \G^k+\alpha^k \Delta\G$, {s}et $k\leftarrow k+1$
\UNTIL{Convergence}
\end{algorithmic}
\end{algorithm}
More specifically, for  problem  \eqref{eq:optreg_disc} we can set 
$$
f(\G)=\left < \G, \C \right >_F  +\eta \Omega_c(\G) \quad \text{and} \quad g(\G)=\lambda\Omega_s(\G).$$ 
Supposing now that $ \Omega_c(\G)$ is differentiable, step 3 of Algorithm
 \ref{algo:fws} boils down to 
    \begin{equation*}
      \G^\star=\argmin_{\G \in \Pp} \quad\left < \G, \C + \eta \nabla\Omega_c(\G^k) \right >_F+ \lambda\Omega_s(\G)
    \end{equation*}
    Interestingly, this problem is an entropy-regularized optimal
    transport problem similar to Problem~\eqref{eq:cuturi}
    and can be efficiently solved using the Sinkhorn-Knopp
    scaling matrix approach.

In our optimal transport problem, 
$\Omega_c(\G)$ is instantiated by the Laplacian 
or  the group-lasso regularization term. 
The former is differentiable whereas the group-lasso is not  when 
there exists a class ${cl}$ and an index $j$ for which
$\G(\mathcal{I}_{cl},j)$ is a vector of $0$. However, 
one can note that if the iterate $\G^k$ is so that  $\G^k(\mathcal{I}_{cl},j) \neq 0
\,\, \forall {cl}, \forall j$, then  the same property holds for
$\G^{k+1}$. This is due  to the exponentiation occurring in the
Sinkhorn-Knopp algorithm
used for the entropy-regularized optimal transport problem. This
means that if we initialize 
 $\G^0$  so that  $\G^0(\mathcal{I}_{cl},j) \neq 0$, then  
$\Omega_c(\G^k)$ is always differentiable. 
Hence, our GCG algorithm can also be applied to the group-lasso regularization,
despite its non-differentiability in $0$.

%%% End: 

 \section{Numerical experiments}
 \label{sec:numer-exper}

In this section, we study the behavior of four different versions of optimal transport applied to  DA problem. In the rest of the section, \textbf{OT-exact} is  the original transport problem~\eqref{eq:gzero}, \textbf{OT-IT}  the Information theoretic regularized one~\eqref{eq:cuturi}, and the two proposed class-based regularized ones are denoted \textbf{OT-GL}  and \textbf{OT-Laplace},  
corresponding respectively to the group-lasso (Equation~\eqref{eq:cuturigroup}) and Laplacian (Equation~\eqref{eq:reglapsurce}) regularization terms. \corv{We also present some results with our previous class-label based regularizer built upon  an $\ell_p-\ell_1$ norm: \textbf{OT-LpL1}~\cite{courty2014domain}}.

\subsection{Two moons: simulated problem with controllable complexity}
\label{sec:inc_complexity}
In the first experiment, we consider the same toy example as  
in~\cite{germain13}. The simulated dataset consists of
two domains: for the source, the standard two entangled moons data,
where each moon is associated to a specific class (See Figure \ref{fig:2moons}(a)). The target
domain is built by applying a rotation to the two moons, which allows
to consider an adaptation problem with an increasing difficulty as a
function of the rotation angle. This example is notably interesting
because the corresponding problem
is clearly non-linear, and because the input dimensionality is small,
$2$, which leads to poor performances when applying methods based on subspace alignment ({\em e.g.}~\cite{gong12,long13}). 

We follow the same experimental protocol as in~\cite{germain13}, thus allowing
 for a direct comparison with the state-of-the-art results presented therein. The source domain is composed of two moons of $150$ samples each. The target domain is also sampled from these two shapes, with the same number of 
examples. Then, the generalization capability of our method is tested over a set of $1000$ samples that follow the same distribution as the target domain. The experiments 
are conducted $10$ times, and we consider the mean classification error as comparison criterion. As a classifier, we used a SVM
with a Gaussian kernel, whose parameters were set by 5-fold cross-validation. We compare the adaptation results with two state-of-the-art methods: the DA-SVM approach~\cite{bruzzone10} and the more recent PBDA~\cite{germain13}, which has proved to provide competitive results over this dataset. 

Results are reported in Table~\ref{tab:results_twomoons}. Our first observation is that all the methods based on optimal
transport behave better than the state-of-the-art methods, in
particular for low rotation angles, where results indicate that the
geometrical structure is better preserved through the adaptation by
optimal transport. Also, for large angle (\eg~$90^{\circ}$), the final
score is also significantly better than other state-of-the-art method,
but falls down to a $0.5$ error rate, which is natural since in this
configuration a transformation of $-90^{\circ}$, implying an inversion of
labels, would have led to similar empirical distributions. This
clearly shows the capacity of our method to handle large domain
transformations. Adding the class-label information into the
regularization also clearly helps for the mid-range angle values,
where  the adaptation shows nearly optimal results up to
angles $<40^{\circ}$. For
the strongest deformation ($>70^\circ$ rotation), no clear winner
among the OT methods can be found. We think that, regardless of the
amount and type of regularization chosen, the classification of test
samples becomes too much tributary of the training samples.
These ones mostly come from the denser part of $\mu_s$ and  as a
consequence, the less dense parts of this PDF are not satisfactorily
transported. This behavior can be seen in Figure~\ref{fig:2moons}d.

 \begin{table}[tp]
 	\begin{center}	
		 \resizebox{\columnwidth}{!}{
		\begin{tabular}{|c|ccccccc|}
		\hline
		 Target rotation angle  & $10^{\circ}$& $20^{\circ}$ & $30^{\circ}$ & $40^{\circ}$ & $50^{\circ}$ & $70^{\circ}$ &  $90^{\circ}$\\
		\hline
		\hline
		SVM (no adapt.)	& $0$ & $0.104$ & $0.24$ & $0.312$ & $0.4$ & $0.764 $ &  $0.828$\\
		DASVM~\cite{bruzzone10} & $0$ & $0$ & $0.259$ & $0.284$ & $0.334$  & $0.747$ &  $0.82$\\
		PBDA~\cite{germain13}	& $0$ & $0.094$ & $0.103$ & $0.225$ & $0.412$  & $0.626$ & $0.687$\\		
		\hline
		\textbf{OT-exact}        & 0 & 0.028 & 0.065 & 0.109 & 0.206 & 0.394 & 0.507 \\
		\textbf{OT-IT}           & 0 & 0.007 & 0.054 & 0.102 & 0.221 & 0.398 & 0.508 \\
		\textbf{OT-GL}           & 0 & 0 & 0 & 0.013 & 0.196 & 0.378 & 0.508 \\
		\textbf{OT-Laplace}      & 0 & 0 & 0.004 & 0.062 & 0.201 & 0.402 & 0.524 \\
		\hline
		\end{tabular}
		}
	\end{center}
	\caption{Mean error rate over $10$ realizations for the two moons simulated example.}\label{tab:results_twomoons}
\end{table} 

\begin{figure*}[!t]
  \centering
  \subfigure[source domain]{\includegraphics[width=0.24\linewidth]{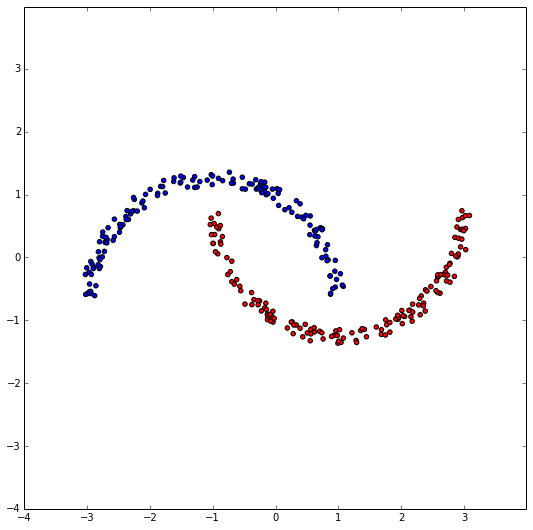}}
  \subfigure[rotation=$20^{\circ}$]{\includegraphics[width=0.24\linewidth]{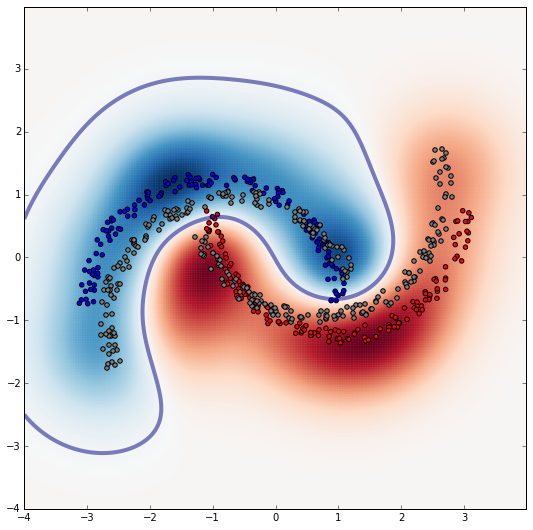}}
  \subfigure[rotation=$40^{\circ}$]{\includegraphics[width=0.24\linewidth]{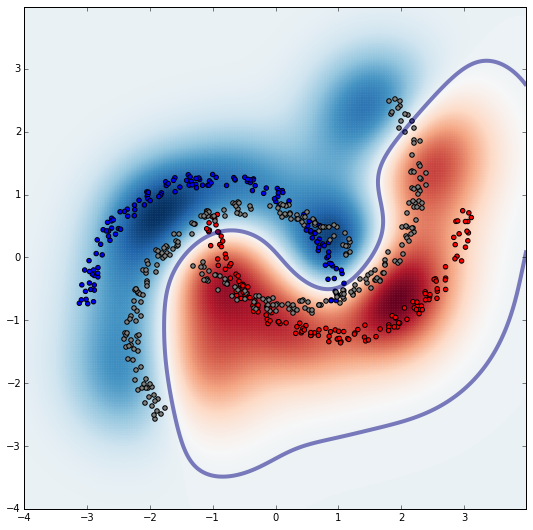}}
  \subfigure[rotation=$90^{\circ}$]{\includegraphics[width=0.24\linewidth]{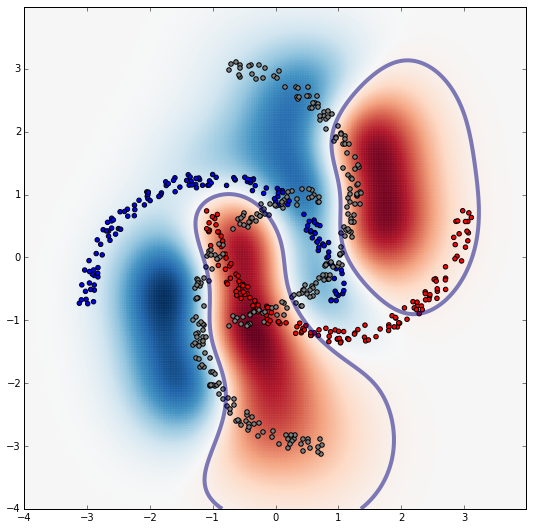}}
  \caption{Illustration of the classification decision boundary produced by \textbf{OT-Laplace} over the two moons example for increasing rotation angles. The source domain is represented as coloured points. The target domain is depicted as points in grey (best viewed with colors).}
  \label{fig:2moons}
\end{figure*}

\subsection{Visual adaptation datasets}
\label{sec:comp-visi-datas}

We now evaluate our method on {three} challenging real world vision
adaptation task{s}, which ha{ve} attracted a lot of interest {in recent computer vision literature}~\cite{Pat14}. We start by presenting the datasets, then the
experimental protocol, {and finish by providing and discussing} the
 results obtained.

\subsubsection{Datasets}
Three types of image {recognition problems} are considered: digits, faces
and miscellaneous objects {recognition}. This choice of datasets was
already featured
in~\cite{long13}. A
summary of the  properties of each domain considered in the three problems
is {provided} in
Table~\ref{tab:datasets}.
An illustration {of some examples} of the different domains for a particular class is shown in Figure~\ref{fig:illus_dataset}.

\noindent{\bf Digit recognition.} {As source and target domains, w}e
use the two digits datasets USPS and MNIST, that share 10 classes of
digits {(single digits $0-9$)}. We randomly sampled $1,800$ and
$2,000$ images from {each} original dataset. The MNIST images
are resized to the same resolution as that of USPS ($16\times16$). The grey levels of all images are then  normalized to obtain a final common feature space for both domains.~\\    
\noindent{\bf Face recognition.} In the face recognition experiment,
we use the PIE ("Pose, Illumination, Expression") dataset, which
contains $32\times32$ images of 68 individuals taken under various
pose, illumination and expressions conditions. The 4 experimental
domains are constructed by selecting 4  distinct poses: PIE05 (C05,
left pose), PIE07 (C07, upward pose), PIE09 (C09, downward pose)
 and PIE29 (C29, right pose). This allows to
define 12 different adaptation problems with increasing difficulty
(the most challenging being the adaptation from right to left
poses). Let us note that each domain {has a }
strong variability for each class {due to} illumination and
expression {variations}.~\\ 
\noindent{\bf Object recognition.} We used the Caltech-Office dataset~\cite{saenko10,gopalan11,gong12,zheng12,Pat14}. The dataset contains images coming from four different domains: {\em Amazon} (online merchant), the {\em Caltech-256} image collection~\cite{griffin07}, {\em Webcam} (images taken from a webcam) and {\em DSLR} (images taken from a high resolution digital SLR camera). The variability of the different domains come from several factors: presence/absence of background, lightning conditions, noise, etc.   \corv{We consider two feature sets:
\begin{itemize}
\item       
  SURF descriptors as described in~\cite{saenko10}, used to transform
  each image into a 800 bins histogram{. These histograms are}
  subsequently normalized and reduced to standard scores.
\item two DeCAF deep learning features sets~\cite{donahue2014decaf}: these features are extracted as the sparse activation of the neurons from the fully connected 6th and 7th layers of a convolutional network trained on imageNet and then fine tuned on the visual recognition tasks considered here. As such, they form vectors with 4096 dimensions. 
\end{itemize}

 \begin{table}[tp]
 	\begin{center}	
		 \resizebox{\columnwidth}{!}{
		\begin{tabular}{|cc|ccccc|}
		\hline
		{\bf Problem} & {\bf Domains} & {\bf Dataset} & {\bf \# Samples} & {\bf \# Features} & {\bf \# Classes} & {\bf Abbr.} \\ 
		\hline
		\hline
		\multirow{2}{*}{Digits}&USPS & USPS & 1800 & 256 & 10 & U \\
		&MNIST & MNIST & 2000 & 256 & 10 & M \\
		\hline
		\multirow{5}{*}{Faces}&PIE05 & PIE & 3332 & 1024 & 68 & P1 \\
		&PIE07 & PIE & 1629 & 1024 & 68 & P2 \\
		&PIE09 & PIE & 1632 & 1024 & 68 & P3 \\
		&PIE29 & PIE & 1632 & 1024 & 68 & P4 \\
		\hline
		\multirow{4}{*}{Objects}&Calltech & Calltech & 1123 & 800$\mid$4096 & 10 & C\\
		&Amazon & Office & 958 & 800$\mid$4096 & 10 & A\\
		&Webcam & Office & 295 & 800$\mid$4096 & 10 & W\\
		&DSLR & Office & 157 & 800$\mid$4096 & 10 & D\\
		\hline
		\end{tabular}
		}
	\end{center}
	\caption{Summary of the domains used in the visual adaptation experiment}\label{tab:datasets}
\end{table}
}

\begin{figure}[!t]
  \centering
  \includegraphics[width=1\columnwidth]{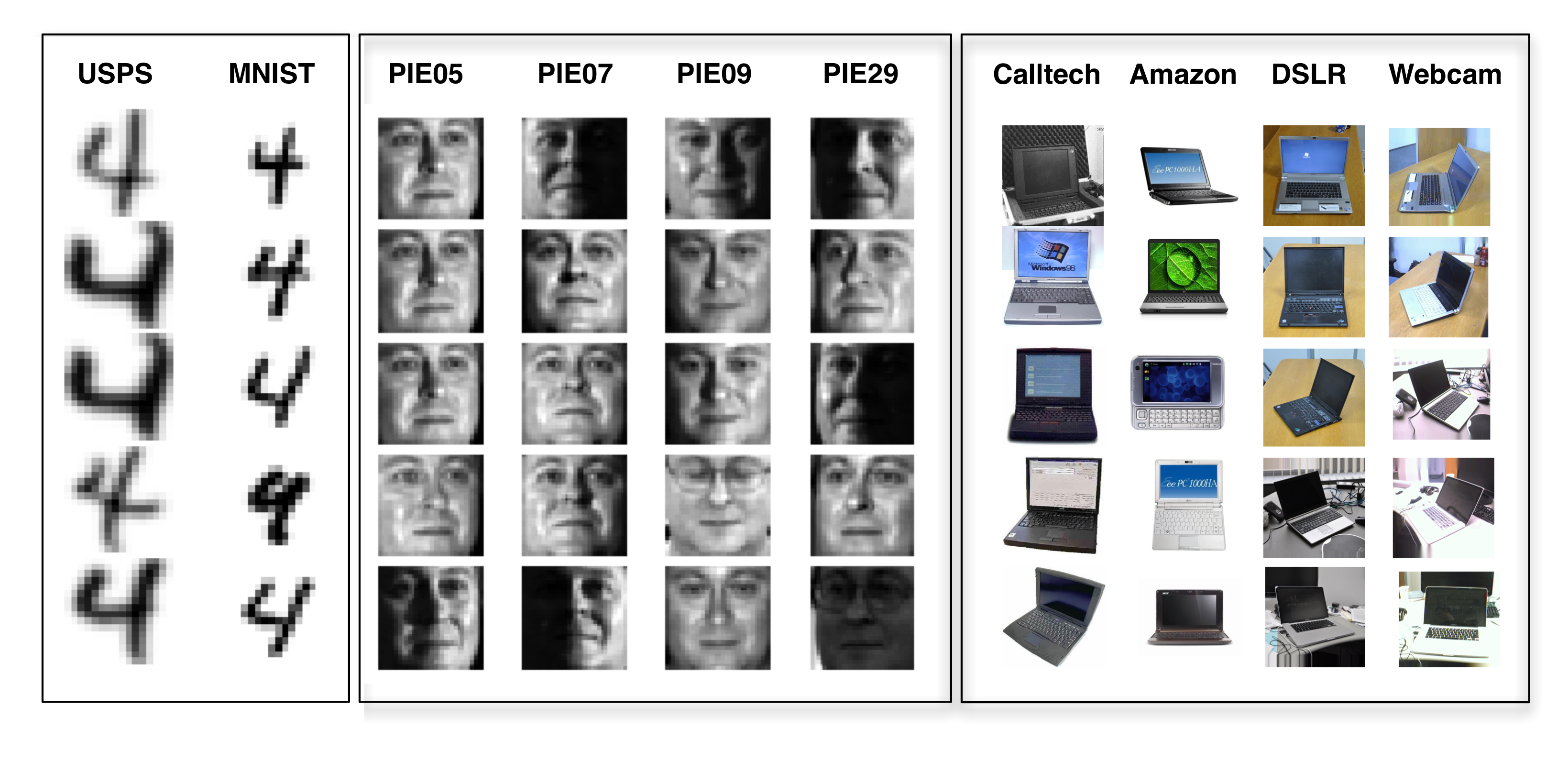}  
	\caption{Examples from the datasets used in the visual adaptation experiment. 5 random samples from one class are given for all the considered domains.}
  \label{fig:illus_dataset}
\end{figure}

\subsubsection{Experimental setup}
\corv{
Following~\cite{gong12}, the classification is conducted using a $1$-Nearest Neighbor ({1}NN) classifier, which has the advantage of being parameter free. In all 
experiments, {1}NN is trained with the adapted source data, and evaluated over the target data to provide a  classification accuracy score.  
We compare our optimal transport solutions to the following baseline methods that are particularly well adapted for image classification:
\begin{itemize}
\item {\bf {1}NN} is the original classifier without adaptation and constitutes a baseline for all experiments;
\item {\bf PCA}, which consists in applying a projection on the first
  principal components of the joint source/target distribution
  (estimated from the concatenation of source and target samples); 
\item {\bf GFK}, Geodesic Flow Kernel~\cite{gong12}; 
\item {\bf TSL}, Transfer Subspace Learning~\cite{Si10}, which operates by minimizing the Bregman divergence between the domains embedded in lower dimensional spaces;
\item {\bf JDA}, Joint Distribution Adaptation~\cite{long13}, which extends the Transfer Component Analysis algorithm~\cite{Pan11};
\end{itemize}

In {unsupervised DA} no target labels are available. As a consequence, it is impossible to 
consider a cross-validation step for the hyper-parameters of the different methods. 
However, and in order to compare the methods fairly, we follow the following protocol. For each source domain, a random selection of 
20 samples per class (with the only exception of 8 for the DSLR dataset) is adopted. Then the target domain is equivalently partitioned in  
a validation and test sets. The validation set is used to obtain the best accuracy in the range of the possible hyper-parameters. The accuracy,
measured as the percent of correct classification over all the classes, is then evaluated on the testing set, with the best selected hyper-parameters. 
This strategy normally prevents overfitting on the testing set. The experimentation is conducted $10$ times, and the mean accuracy over all these
realizations is reported.

We considered the following parameter range : for
 subspace learning methods ({\bf PCA},{\bf TSL}, {\bf GFK},  and {\bf JDA}) we considered reduced $k$-dimensional spaces with $k\in\{10,20,\hdots,70\}$. A linear kernel was chosen for all 
 the methods with a kernel formulation. For the all methods requiring a regularization parameter, the best value was searched in $\lambda = \{0.001,0.01,0.1,1,10,100,1000\}$. The $\lambda$ and $\eta$ parameters of our different regularizers
(Equation~\eqref{eq:optreg_disc}), are validated using the same search interval. In the case of the Laplacian regularization (\textbf{OT-Laplace}), 
$\mathbf{S}_{t}$ is a binary matrix which encodes a nearest neighbors graph with a 8-connectivity.  For the source domain, $\mathbf{S}_s$ is filtered such that connections between elements of different
classes are pruned. Finally, we set the $\alpha$ value Equation~\eqref{eq:reglaps2} to $0.5$.
}

\subsubsection{Results on unsupervised domain adaptation}

\begin{table*}[!tb]
\caption{Overall recognition accuracies in $\%$ obtained over all domains pairs using the SURF features. Maximum values for each pair is indicated in bold font.}
\begin{center}
{
		
	\label{tab:table}
		\begin{tabular}{c|cc|ccc|ccccc}
		\toprule
			{Domains} & {{\bf 1NN}} & {{\bf PCA}} & {{\bf GFK}} & {{\bf TSL}} & {{\bf JDA}} & {{\bf OT-exact}} & {{\bf OT-IT}} & {{\bf OT-Laplace}} & {{\bf OT-LpLq}} & {{\bf OT-GL}}\\
			\midrule
			U$\rightarrow$M   & $39.00$ & $37.83$ & $44.16$ & $40.66$ & $54.52$ & $50.67$ & $53.66$ & $57.42$ & $\bf 60.15$ & $57.85$\\
			M$\rightarrow$U   & $58.33$ & $48.05$ & $60.96$ & $53.79$ & $60.09$ & $49.26$ & $64.73$ & $64.72$ & $68.07$ & $\bf 69.96$\\\hdashline
			mean & $48.66$ & $42.94$ & $52.56$ & $47.22$ & $57.30$ & $49.96$ & $59.20$ & $61.07$ & $\bf 64.11$ & $63.90$\\ \hline
			P1$\rightarrow$P2 & $23.79$ & $32.61$ & $22.83$ & $34.29$ & $\bf 67.15$ & $52.27$ & $57.73$ & $58.92$ & $59.28$ & $59.41$\\
			P1$\rightarrow$P3 & $23.50$ & $38.96$ & $23.24$ & $33.53$ & $56.96$ & $51.36$ & $57.43$ & $57.62$ & $58.49$ & $\bf 58.73$\\
			P1$\rightarrow$P4 & $15.69$ & $30.82$ & $16.73$ & $26.85$ & $40.44$ & $40.53$ & $47.21$ & $47.54$ & $47.29$ & $\bf 48.36$\\
			P2$\rightarrow$P1 & $24.27$ & $35.69$ & $24.18$ & $33.73$ & $\bf 63.73$ & $56.05$ & $60.21$ & $62.74$ & $62.61$ & $61.91$\\
			P2$\rightarrow$P3 & $44.45$ & $40.87$ & $44.03$ & $38.35$ & $\bf 68.42$ & $59.15$ & $63.24$ & $64.29$ & $62.71$ & $64.36$\\
			P2$\rightarrow$P4 & $25.86$ & $29.83$ & $25.49$ & $26.21$ & $49.85$ & $46.73$ & $51.48$ & $\bf 53.52$ & $50.42$ & $52.68$\\
			P3$\rightarrow$P1 & $20.95$ & $32.01$ & $20.79$ & $39.79$ & $\bf 60.88$ & $54.24$ & $57.50$ & $57.87$ & $58.96$ & $57.91$\\
			P3$\rightarrow$P2 & $40.17$ & $38.09$ & $40.70$ & $39.17$ & $65.07$ & $59.08$ & $63.61$ & $\bf 65.75$ & $64.04$ & $64.67$\\
			P3$\rightarrow$P4 & $26.16$ & $36.65$ & $25.91$ & $36.88$ & $52.44$ & $48.25$ & $52.33$ & $\bf 54.02$ & $52.81$ & $52.83$\\
			P4$\rightarrow$P1 & $18.14$ & $29.82$ & $20.11$ & $40.81$ & $\bf 46.91$ & $43.21$ & $45.15$ & $45.67$ & $46.51$ & $45.73$\\
			P4$\rightarrow$P2 & $24.37$ & $29.47$ & $23.34$ & $37.50$ & $\bf 55.12$ & $46.76$ & $50.71$ & $52.50$ & $50.90$ & $51.31$\\
			P4$\rightarrow$P3 & $27.30$ & $39.74$ & $26.42$ & $46.14$ & $\bf 53.33$ & $48.05$ & $52.10$ & $52.71$ & $51.37$ & $52.60$\\\hdashline
			mean & $26.22$ & $34.55$ & $26.15$ & $36.10$ & $\bf 56.69$ & $50.47$ & $54.89$ & $56.10$ & $55.45$ & $55.88$\\ \hline
			C$\rightarrow$A   & $20.54$ & $35.17$ & $35.29$ & $45.25$ & $40.73$ & $30.54$ & $37.75$ & $38.96$ & $\bf 48.21$ & $44.17$\\
			C$\rightarrow$W   & $18.94$ & $28.48$ & $31.72$ & $37.35$ & $33.44$ & $23.77$ & $31.32$ & $31.13$ & $38.61$ & $\bf 38.94$\\
			C$\rightarrow$D   & $19.62$ & $33.75$ & $35.62$ & $39.25$ & $39.75$ & $26.62$ & $34.50$ & $36.88$ & $39.62$ & $\bf 44.50$\\
			A$\rightarrow$C   & $22.25$ & $32.78$ & $32.87$ & $\bf 38.46$ & $33.99$ & $29.43$ & $31.65$ & $33.12$ & $35.99$ & $34.57$\\
			A$\rightarrow$W   & $23.51$ & $29.34$ & $32.05$ & $35.70$ & $36.03$ & $25.56$ & $30.40$ & $30.33$ & $35.63$ & $\bf 37.02$\\
			A$\rightarrow$D   & $20.38$ & $26.88$ & $30.12$ & $32.62$ & $32.62$ & $25.50$ & $27.88$ & $27.75$ & $36.38$ & $\bf 38.88$\\
			W$\rightarrow$C   & $19.29$ & $26.95$ & $27.75$ & $29.02$ & $31.81$ & $25.87$ & $31.63$ & $31.37$ & $33.44$ & $\bf 35.98$\\
			W$\rightarrow$A   & $23.19$ & $28.92$ & $33.35$ & $34.94$ & $31.48$ & $27.40$ & $37.79$ & $37.17$ & $37.33$ & $\bf 39.35$\\
			W$\rightarrow$D   & $53.62$ & $79.75$ & $79.25$ & $80.50$ & $\bf 84.25$ & $76.50$ & $80.00$ & $80.62$ & $81.38$ & $84.00$\\
			D$\rightarrow$C   & $23.97$ & $29.72$ & $29.50$ & $31.03$ & $29.84$ & $27.30$ & $29.88$ & $31.10$ & $31.65$ & $\bf 32.38$\\
			D$\rightarrow$A   & $27.10$ & $30.67$ & $32.98$ & $36.67$ & $32.85$ & $29.08$ & $32.77$ & $33.06$ & $37.06$ & $\bf 37.17$\\
			D$\rightarrow$W   & $51.26$ & $71.79$ & $69.67$ & $77.48$ & $80.00$ & $65.70$ & $72.52$ & $76.16$ & $74.97$ & $\bf 81.06$\\\hdashline
			mean & $28.47$ & $37.98$ & $39.21$ & $42.97$ & $44.34$ & $36.69$ & $42.30$ & $43.20$ & $46.42$ & $\bf 47.70$\\ 
			\bottomrule
		\end{tabular}		
}
\label{tab:realdata}
\end{center}
\end{table*}

Results of the experiment are reported in Table~\ref{tab:realdata} where the best performing method for each domain adaptation problem is
highlighted in bold.  On average, all the OT-based domain  adaptation methods {perform} {better than} the baseline methods, except in the case of the PIE dataset, where 
 {\bf JDA} outperforms the OT-based methods in 7 out of 12 domain pairs. A possible explanation is that the 
 dataset contains a lot of classes (68), and the EM-like step of {\bf JDA}, which allows to take into account the current 
 results of classification on the target, is clearly leading to a benefit\DT{, except on 5 domain pairs (out of 12), where {\bf OT-Laplace} and 
 {\bf OT-GL} are performing best}{}. We notice that {\bf TSL}, which is based on a similar principle of distribution divergence minimization, almost never outperforms our regularized strategies, except on pair A$\rightarrow$C.
 Among the different optimal transport strategies,  {\bf OT-Exact} leads to the lowest performances. {\bf OT-IT}, the entropy regularized version 
of the transport, is substantially better than {\bf OT-Exact}, but is still inferior to the class-based regularized strategies proposed in this paper.
 The best performing strategies are clearly {\bf OT-GL} and {\bf OT-Laplace} with a slight advantage for {\bf OT-GL}. 
{\bf OT-LpL1}, which is based on a similar regularization strategy as {\bf OT-GL}, but with a different optimization scheme, has globally inferior performances, except on 
some pairs of domains (e.g.  C$\rightarrow$A ) where it achieves better scores. On both digit\DT{}{s} and object\DT{}{s} recognition tasks, {\bf OT-GL}
 significantly outperforms the baseline methods.

\corv{
In the next experiment (Table~\ref{tab:decaf}), we use the same experimental protocol on different features produced by the DeCAF deep learning 
architecture~\cite{donahue2014decaf}. We report the results of the experiment conducted on the Office-Caltech dataset, with 
the {{\bf OT-IT}} and {{\bf OT-GL}} regularization strategies. For comparison purposes, 
{\bf JDA} is also considered for this adaptation task. The results show that, even though the deep learning features yield naturally a strong 
improvement over the classical SURF features,  {the proposed OT methods are}  still capable of improving significantly the performances of the final classification (up to more than 20
points in some case, e.g. D$\rightarrow$A or A$\rightarrow$W). This clearly shows how {OT has the} capacity to handle non-stationarity in the distributions  
that the deep architecture has difficulty handling. We also note that using the features from the 7th  layer instead of the 6th does not bring 
a strong improvement in the classification accuracy, suggesting that part of the work of the 7th layer is already performed by {the optimal transport}. 

\begin{table}[t]
	\caption{Results of adaptation by optimal transport using DeCAF features.}\label{tab:decaf}
	\begin{center}
	\label{tab:table}\resizebox{\columnwidth}{!}{
		\begin{tabular}{c|cccc|cccc}
		\toprule
			{} & \multicolumn{4}{c}{Layer 6} & \multicolumn{4}{c}{Layer 7}\\\cmidrule(r){2-5}\cmidrule(r){6-9}
			{Domains} & {{\bf DeCAF}} & {{\bf JDA}} & {{\bf OT-IT}} & {{\bf OT-GL}} & {{\bf DeCAF}} & {{\bf JDA}} & {{\bf OT-IT}} & {{\bf OT-GL}}\\
			\midrule
			C$\rightarrow$A & $79.25$ & $88.04$ & $88.69$ & $\bf 92.08$ & $85.27$ & $89.63$ & $91.56$ & $\bf 92.15$\\
			C$\rightarrow$W & $48.61$ & $79.60$ & $75.17$ & $\bf 84.17$ & $65.23$ & $79.80$ & $82.19$ & $\bf 83.84$\\
			C$\rightarrow$D & $62.75$ & $84.12$ & $83.38$ & $\bf 87.25$ & $75.38$ & $85.00$ & $85.00$ & $\bf 85.38$\\
			A$\rightarrow$C & $64.66$ & $81.28$ & $81.65$ & $\bf 85.51$ & $72.80$ & $82.59$ & $84.22$ & $\bf 87.16$\\
			A$\rightarrow$W & $51.39$ & $80.33$ & $78.94$ & $\bf 83.05$ & $63.64$ & $83.05$ & $81.52$ & $\bf 84.50$\\
			A$\rightarrow$D & $60.38$ & $\bf 86.25$ & $85.88$ & $85.00$ & $75.25$ & $85.50$ & $\bf 86.62$ & $85.25$\\
			W$\rightarrow$C & $58.17$ & $\bf 81.97$ & $74.80$ & $81.45$ & $69.17$ & $79.84$ & $81.74$ & $\bf 83.71$\\
			W$\rightarrow$A & $61.15$ & $90.19$ & $80.96$ & $\bf 90.62$ & $72.96$ & $90.94$ & $88.31$ & $\bf 91.98$\\
			W$\rightarrow$D & $97.50$ & $\bf 98.88$ & $95.62$ & $96.25$ & $98.50$ & $\bf 98.88$ & $98.38$ & $91.38$\\
			D$\rightarrow$C & $52.13$ & $81.13$ & $77.71$ & $\bf 84.11$ & $65.23$ & $81.21$ & $82.02$ & $\bf 84.93$\\
			D$\rightarrow$A & $60.71$ & $91.31$ & $87.15$ & $\bf 92.31$ & $75.46$ & $91.92$ & $92.15$ & $\bf 92.92$\\
			D$\rightarrow$W & $85.70$ & $\bf 97.48$ & $93.77$ & $96.29$ & $92.25$ & $\bf 97.02$ & $96.62$ & $94.17$\\\hdashline
			mean & $65.20$ & $86.72$ & $83.64$ & $\bf 88.18$ & $75.93$ & $87.11$ & $87.53$ & $\bf 88.11$\\
		\bottomrule
		\end{tabular}}
	\end{center}
\end{table}

}

\corv{
\subsubsection{Semi-supervised domain adaptation}
In this last experiment, we assume that few labels are available in
the target domain. We thus benchmark our semi-supervised 
approach  on SURF features extracted from
the Office-Caltech dataset.  We consider that only  3 labeled
samples per class are at our disposal in the target domain.  In order to disentangle the benefits of the
labeled target samples brought by our optimal transport
strategies from those brought by the classifier, we make a distinction between two
cases: in the first one, denoted as ``Unsupervised + labels'', we consider that the label target samples are available only at
the learning stage, after an unsupervised domain 
adaptation with optimal transport.
 In the second case, denoted as  ``semi-supervised'', labels in the target domain are used to compute a new transportation plan,
 through the use of the proposed {semi-supervised} regularization term
in {Equation~\eqref{eq:ssl}}).

Results are reported in Table
  \ref{tab:table2}. 
They clearly show the benefits of the proposed semi-supervised regularization term in the
definition of the transportation plan. A comparison with the
state-of-the-art method of Hoffman and
colleagues~\cite{Hoffman_ICLR2013} is also reported, and shows the
competitiveness of our approach.    

\begin{table}[t]
	\begin{center}
		\caption{Results of semi-supervised adaptation with optimal transport using the SURF features.}\label{tab:table2}
\resizebox{\columnwidth}{!}{s
		\begin{tabular}{cccccc}
		\toprule
			{} & \multicolumn{2}{c}{Unsupervised + labels} & \multicolumn{3}{c}{Semi-supervised}\\\cmidrule(r){2-3}\cmidrule(r){4-6}
			{Domains} & {{\bf OT-IT}} & {{\bf OT-GL}} & {{\bf OT-IT}} & {{\bf OT-GL}} & {{\bf MMDT}~\cite{Hoffman_ICLR2013}}\\
			\midrule
			C$\rightarrow$A & 37.0 $\pm$ 0.5 & 41.4 $\pm$ 0.5 & 46.9 $\pm$ 3.4 & 47.9 $\pm$ 3.1 & \textbf{49.4 $\pm$ 0.8}\\
			C$\rightarrow$W & 28.5 $\pm$ 0.7 & 37.4 $\pm$ 1.1 & 64.8 $\pm$ 3.0 & \textbf{65.0 $\pm$ 3.1} & 63.8 $\pm$ 1.1\\
			C$\rightarrow$D   & 35.1 $\pm$ 1.7 & 44.0 $\pm$ 1.9 & 59.3 $\pm$ 2.5 & \textbf{61.0 $\pm$ 2.1} & 56.5 $\pm$ 0.9\\
			A$\rightarrow$C & 32.3 $\pm$ 0.1 & 36.7 $\pm$ 0.2 & 36.0 $\pm$ 1.3 & \textbf{37.1 $\pm$ 1.1} & 36.4 $\pm$ 0.8\\
			A$\rightarrow$W    & 29.5 $\pm$ 0.8 & 37.8 $\pm$ 1.1 & 63.7 $\pm$ 2.4 & \textbf{64.6 $\pm$ 1.9} & \textbf{64.6 $\pm$ 1.2}\\
			A$\rightarrow$D      & 36.9 $\pm$ 1.5 & 46.2 $\pm$ 2.0 & 57.6 $\pm$ 2.5 & \textbf{59.1 $\pm$ 2.3} & 56.7 $\pm$ 1.3\\
			W$\rightarrow$C & 35.8 $\pm$ 0.2 & 36.5 $\pm$ 0.2 & 38.4 $\pm$ 1.5 & \textbf{38.8 $\pm$ 1.2} & 32.2 $\pm$ 0.8\\
			W$\rightarrow$A    & 39.6 $\pm$ 0.3 & 41.9 $\pm$ 0.4 & 47.2 $\pm$ 2.5 & 47.3 $\pm$ 2.5 & \textbf{47.7$\pm$ 0.9}\\
			W$\rightarrow$D      & 77.1 $\pm$ 1.8 & \textbf{80.2 $\pm$ 1.6} & 79.0 $\pm$ 2.8 & 79.4 $\pm$ 2.8 & 67.0 $\pm$ 1.1\\
			D$\rightarrow$C   & 32.7 $\pm$ 0.3 & 34.7 $\pm$ 0.3 & 35.5 $\pm$ 2.1 & 3\textbf{6.8 $\pm$ 1.5} & 34.1 $\pm$ 1.5\\
			D$\rightarrow$A      & 34.7 $\pm$ 0.3 & 37.7 $\pm$ 0.3 & 45.8 $\pm$ 2.6 & 46.3 $\pm$ 2.5 & \textbf{46.9 $\pm$ 1.0}\\
			D$\rightarrow$W      & 81.9 $\pm$ 0.6 &  \textbf{84.5 $\pm$ 0.4} & 83.9 $\pm$ 1.4 & 84.0 $\pm$ 1.5 & 74.1 $\pm$ 0.8\\\hdashline
			mean &                     41.8 & 46.6 & 54.8 & \textbf{55.6} & 52.5\\
		\bottomrule
		\end{tabular}
		}
	\end{center}
\end{table}

}

\section{Conclusion}

In this paper, we described a new framework based on optimal
transport to solve the unsupervised domain adaptation problem.
We proposed two regularization schemes to encode class-structure in
the source domain during the estimation of the transportation plan,
thus enforcing the intuition that samples of the same class must
undergo similar transformation. We extended this {OT regularized
  framework} to the semi-supervised domain adaptation case,
i.e. the case where few labels are available in the target domain.
Regarding the computational aspects, we suggested to use a modified
version of the conditional gradient algorithm, the generalized
conditional gradient splitting, which enables the method to
scale up to real-world datasets. Finally, we applied the proposed
methods on both synthetic and real world datasets. Results show that
the optimal transportation domain adaptation schemes frequently
outperform the competing state-of-the-art methods.

We believe that the framework presented in this paper will lead to
a paradigm shift for the domain adaptation problem. Estimating a
transport is much more general than finding a common subspace, but 
comes with the problem of finding a proper regularization term. The proposed class-based or Laplacian regularizers show very good performances, but we believe that 
other types of regularizer should be investigated. Indeed, whenever
the transformation is induced by a physical process, one may want the
transport map to enforce 
physical constraints.  This can be included with dedicated
regularization terms. We also
plan to extend our  optimal transport framework to the multi-domain adaptation problem, where the problem of
matching several distributions can be cast as a multi-marginal optimal transport problem.

\ifCLASSOPTIONcompsoc
    \section*{Acknowledgments}
\else
    \section*{Acknowledgment}
\fi

This work was partly funded by the Swiss National Science Foundation under the grant PP00P2-150593 and by the CNRS PEPS Fascido program under the
Topase project.

\ifCLASSOPTIONcaptionsoff
  \newpage
\fi

\bibliographystyle{IEEEtranS}
{\footnotesize
% Generated by IEEEtranS.bst, version: 1.13 (2008/09/30)

}
\begin{IEEEbiography}[{\includegraphics[width=1in,height=1.25in,clip,keepaspectratio]{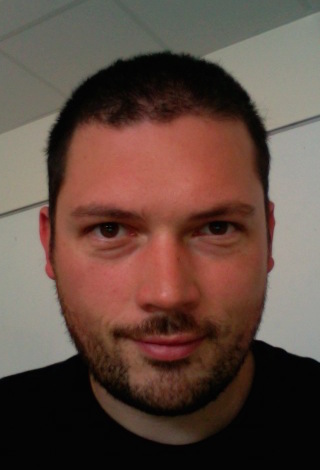}}]{Nicolas Courty} is associate professor within University Bretagne-Sud since October 2004. He obtained his habilitation degree (HDR) in 2013. His main research objectives are data analysis/synthesis schemes, machine learning and visualization problems, with applications in computer vision, remote sensing and computer graphics. 
Visit http://people.irisa.fr/Nicolas.Courty/ for more information.
\end{IEEEbiography}

\begin{IEEEbiography}[{\includegraphics[width=1in,height=1.25in,clip,keepaspectratio]{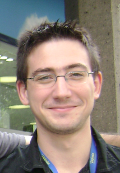}}]{R\'emi Flamary} is Assistant Professor at
Universit\'e C\^ote d'Azur (UCA) and a member of Lagrange
Laboratory/Observatoire de la C\^ote d'Azur since 2012.  He received
a Dipl.-Ing. in electrical engineering and a M.S. degrees in image
processing from the Institut National de Sciences Appliqu\'ees de Lyon
in 2008 and  a Ph.D. degree from the University of Rouen in
2011. His current research interest involve signal processing, machine
learning and image processing. 
\end{IEEEbiography}

\begin{IEEEbiography}[{\includegraphics[width=1in,height=1.25in,clip,keepaspectratio]{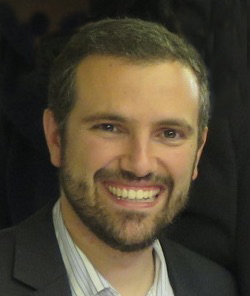}}]{Devis Tuia} (S'07, M'09, SM'15)  received the Ph.D. from University of Lausanne in 2009.
He was a  Postdoc at the University of Val{\'e}ncia, the University of Colorado, Boulder, CO and EPFL Lausanne. Since 2014, he is Assistant Professor with the Department of Geography, University of Zurich. He is interested in algorithms for information extraction and data fusion of remote sensing images using machine learning. 
 More info on http://devis.tuia.googlepages.com/
\end{IEEEbiography}

\begin{IEEEbiography}[{\includegraphics[width=1in,height=1.25in,clip,keepaspectratio]{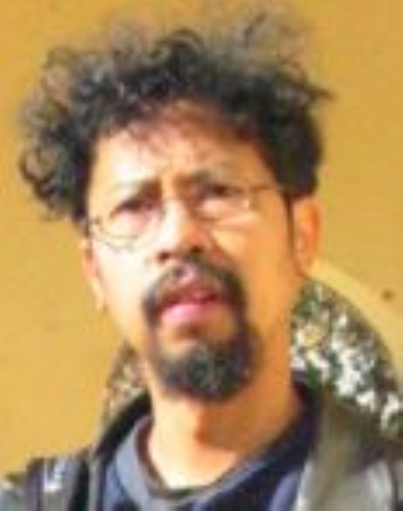}}]{Alain Rakotomamonjy} (M'15)  is Professor in the
Physics department at the University of
Rouen since 2006. He obtained his Phd on
Signal processing from the university of Orl\'eans
in 1997. His recent research activities
deal with machine learning and signal processing
with applications to brain-computer
interfaces and audio applications. Alain serves as a regular
reviewer for machine learning and signal processing journals.
\end{IEEEbiography}

\end{document}